\documentclass{article} 
\usepackage{iclr2026_conference,times}

\usepackage{amsmath,amsfonts,bm}

\def\eqref#1{equation~\ref{#1}}

\def\1{\bm{1}}

\DeclareMathAlphabet{\mathsfit}{\encodingdefault}{\sfdefault}{m}{sl}
\SetMathAlphabet{\mathsfit}{bold}{\encodingdefault}{\sfdefault}{bx}{n}

\usepackage{hyperref}
\usepackage{url}
\usepackage{booktabs}
\usepackage{multirow}
\usepackage{colortbl}    
\usepackage{xcolor}
\usepackage{amsmath}
\usepackage{amssymb}
\usepackage{graphicx}
\usepackage{subcaption}
\usepackage{microtype}      
\usepackage{booktabs}
\usepackage{listings}       

\lstdefinestyle{promptbox}{
  basicstyle=\ttfamily\scriptsize,
  breaklines=true,
  breakindent=0pt,
  columns=fullflexible,
  keepspaces=true,
  showstringspaces=false,
  frame=single,
  rulecolor=\color{gray!50},
  backgroundcolor=\color{gray!5},
  xleftmargin=4pt,
  xrightmargin=4pt,
  framexleftmargin=2pt,
  aboveskip=6pt,
  belowskip=6pt,
}

\definecolor{gaingreen}{RGB}{0,128,0}
\newcommand{\gain}[1]{{\scriptsize\,\textcolor{gaingreen}{($\uparrow$#1)}}}

\definecolor{oursblue}{RGB}{222,235,247}
\definecolor{groupgray}{RGB}{242,242,242}

\title{BrowserForge: Scaling Web Episode \\ via Parallel Browser Sandboxes}

\author{\textbf{Fei Tang}\textsuperscript{1,2},
\textbf{Huawen Shen}\textsuperscript{2},
\textbf{Zhiqiong Lu}\textsuperscript{2},
\textbf{Zhengxi Lu}\textsuperscript{1},
\textbf{Pengyuan Lyu}\textsuperscript{2},
\textbf{Chengquan Zhang}\textsuperscript{2,$\ddagger$},\\
\textbf{Weiming Lu}\textsuperscript{1},
\textbf{Jun Xiao}\textsuperscript{1},
\textbf{Yueting Zhuang}\textsuperscript{1},
\textbf{Yongliang Shen}\textsuperscript{1,$*$},
\\[4pt]
\normalfont
\textsuperscript{1}Zhejiang University \quad
\textsuperscript{2}LLM Department, Tencent\\[2pt]
\textsuperscript{$*$}Corresponding author. \quad
\textsuperscript{$\ddagger$}Project leader.
}

\iclrfinalcopy 
\begin{document}

\fancypagestyle{firstpage}{%
  \fancyhf{}%
  \renewcommand{\headrulewidth}{0.4pt}%
  \lhead{\raisebox{-0.1\height}{%
    \includegraphics[height=0.68cm]{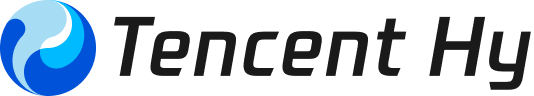}}}%
}

\maketitle

\fancyhead{}
\thispagestyle{firstpage}

\begin{abstract}
Web agents that act from rendered pixels avoid the fragility and heavy
token cost of reading a page's HTML or accessibility tree, but training them
depends on large amounts of high-quality interaction trajectories, and how to
produce such data at scale remains an open problem. Public datasets typically
contain only a few thousand trajectories drawn from a fixed and narrow set of
websites, and even recent automated synthesis pipelines stay bound to predefined
site lists or tutorial sources, so the number of \emph{distinct} websites the
agent ever sees barely grows. We present \textbf{BrowserForge}, a framework that
generates web interaction data at scale by driving many browser sandboxes in
parallel over the open web. BrowserForge couples three components: an open-web
sourcing stage that exposes the agent to hundreds of thousands of real, openly
reachable websites; a sandbox cluster manager that schedules hundreds of concurrent
browsers with high utilization; and a Proposer--Solver dual-agent loop that turns
a raw page into an executable task and then collects a verified trajectory for it.
A rule-plus-model cleaning pipeline removes failed runs and rewrites the surviving
reasoning into a single unified chain-of-thought style. Page structure such as the
accessibility tree is used only as a synthesis-time signal; the agent we train and
release acts purely from the screenshot. The resulting corpus contains $203{,}238$
trajectories, each collected from a distinct website, larger and more diverse than
prior trajectory datasets. Fine-tuning a compact multimodal model on this corpus raises
its success rate on the live \textit{Online-Mind2Web} from $25.66\%$ to
$33.33\%$ and consistently improves step accuracy on the static
\textit{Multimodal-Mind2Web}, with the gain growing as the corpus
scales. 
Controlled analyses further confirm that open-web sourcing and broad website coverage are key contributors to the observed improvement.
\end{abstract}

\section{Introduction}
\label{sec:intro}

Web agents automate tasks on websites for people: given a goal in natural
language, an agent perceives a web page, decides where to act, and carries the
goal across many steps of clicking, typing, and navigating until the task is
done~\citep{wei2025webagent, xiao2026webworld, logeswaran2026scaling, gur2024real,deng2023mind2web,zhou2023webarena,he2024webvoyager,wu2024osatlas}. As more of daily
work moves into the browser, an agent that can operate arbitrary websites on a
user's behalf has become a central goal for multimodal models.

How an agent perceives a page is a design choice. One line of work reads the page
as text, namely its HTML or accessibility (a11y) tree, and acts on element
identifiers~\citep{zheng2024seeact, deng2023mind2web, zhou2023webarena, lu2024weblinx, tang2025survey}; another
operates the page from rendered pixels and grounds its actions in screen
coordinates, the way a person
does~\citep{hong2024cogagent,wu2024osatlas,zheng2024seeact}. The textual route
has two liabilities. It is \emph{fragile}: HTML structure and a11y labels differ
across sites and shift whenever a page is restyled, so a policy bound to them
transfers poorly across the open web~\citep{gou2025uground,lai2024autowebglm}. And it is \emph{costly}: a single page's
a11y tree routinely serializes to tens of thousands of tokens, inflating context
length at every step~\citep{cheng2024seeclick,gou2025uground}. A pure-GUI agent avoids both. A screenshot carries the
page at a fixed, far smaller cost, in a form uniform across sites and aligned with
how interfaces are built to be used~\citep{gupta2026molmoweb,guilibra}.

Training a capable pure-GUI agent, however, demands interaction trajectories at
scale, and crucially over \emph{many distinct websites}, since what the agent
learns is bounded by the layouts and interaction patterns it has seen. Public
datasets fall short: they contain only a few thousand human demonstrations over
tens to low hundreds of sites~\citep{deng2023mind2web,lu2024weblinx}. Automated
synthesis pipelines multiply the trajectory \emph{count}, but each stays bound to
the source it starts from, whether a tutorial corpus~\citep{xu2025agenttrek}, a
curated seed list~\citep{pahuja2025explorer}, or a fixed
environment~\citep{murty2024nnetnav,he2024openwebvoyager}, so the agent still
sees the same narrow set of sites. The scale that matters for the open web, the
number of distinct websites, barely grows.

We build on a simple insight: the open web is itself the largest and cheapest
source of \emph{distinct} websites, and a browser sandbox is a cheap, disposable
unit of \emph{interaction} with any of them. Run many sandboxes in parallel over
openly sourced URLs rather than a fixed list, and scale and diversity grow
together, provided the trajectories are verified and cleaned. 

On this insight we propose \textbf{BrowserForge}, a framework that turns the open web into pure-GUI
web-agent training data through three components: a \emph{sourcing and cleaning}
stage that draws URLs from Common Crawl~\citep{commoncrawl} and filters them down
to hundreds of thousands of real, reachable websites; a \emph{sandbox cluster manager} that
schedules hundreds of browsers through a shared work queue at high utilization;
and a \emph{Proposer--Solver} loop that turns each page into an executable task
and records a trajectory, followed by rule- and model-based cleaning that rewrites
the surviving reasoning into a unified chain-of-thought format. Page structure
such as the a11y tree is used only as a synthesis-time signal; the agent we train
and release acts purely from the screenshot, so none of it enters inference.

Running this pipeline produces $203{,}238$ trajectories, each collected from a
distinct website, larger in count and far broader in coverage than prior trajectory
datasets (Table~\ref{tab:dataset_compare}). We experiment on two web-agent
benchmarks: the live \textit{Online-Mind2Web}~\citep{xue2025illusion} and the static \textit{Multimodal-Mind2Web}.
Fine-tuning a compact backbone on our corpus lifts its Online-Mind2Web success
rate from $25.66\%$ to $33.33\%$ ($+7.67$) and consistently improves
Multimodal-Mind2Web step accuracy, reaching a level competitive with much larger
open-source web agents. Controlled analyses attribute the gain to the open-web
data source and its website diversity rather than to the training recipe, and
show performance scaling monotonically with corpus size; ablations confirm that
each part of the verification-and-cleaning pipeline contributes. Together these
results show that BrowserForge is an effective recipe for producing pure-GUI
web-agent data at scale. 

We make three contributions:
\begin{itemize}
  \item We present \textbf{BrowserForge}, a data-generation framework that scales
  web interaction trajectories by orchestrating parallel browser sandboxes over
  openly sourced URLs, decoupling data scale and diversity from any fixed
  website list.
  \item We design a \textbf{Proposer--Solver} synthesis loop with an explicit
  cleaning and verification pipeline that turns unsupervised open-web pages into
  executable tasks and filtered trajectories with a unified reasoning format.
  \item We curate a corpus of \textbf{$203{,}238$ trajectories, each from a
  distinct website}, and show that fine-tuning compact multimodal models on it
  raises Online-Mind2Web success rate to $33.33\%$ and Multimodal-Mind2Web
  average step accuracy to $43.8\%$, competitive with much larger open-source
  web agents.
\end{itemize}

\section{Related Work}
\label{sec:related}

\paragraph{Automated trajectory synthesis.}
Because human annotation of multi-step trajectories is slow and costly, recent
work synthesizes trajectories automatically, and this line is the closest to
ours. A first group converts existing human-written knowledge into supervision:
AgentTrek~\citep{xu2025agenttrek} harvests tutorial text from the web, turns it
into task specifications, and replays the steps with a vision-language agent
while a model verifier checks correctness, and
Synatra~\citep{ou2024synatra} likewise re-purposes indirect knowledge such as
online tutorials into direct demonstrations at scale. A second group generates
trajectories through interaction: NNetNav~\citep{murty2024nnetnav} explores
environments without supervision and relabels the results retroactively;
OpenWebVoyager~\citep{he2024openwebvoyager} alternates imitation learning with
exploration--feedback cycles; OS-Genesis~\citep{sun2025osgenesis} reverses the
usual pipeline by acting first and deriving tasks afterward (reverse task
synthesis); and Learn-by-interact~\citep{su2025learnbyinteract} synthesizes
agent--environment interactions from documentation and reconstructs the matching
instructions. Explorer~\citep{pahuja2025explorer} builds a bottom-up multi-agent
pipeline that explores websites dynamically and produces over $94$K
trajectories, while GUICourse/GUIAct~\citep{chen2024guicourse} contributes large
GUI navigation datasets for training VLM agents. These methods substantially
increase trajectory counts and report strong downstream gains. Their coverage,
however, is still anchored to the source they start from: a tutorial corpus, a
curated seed list, or a fixed environment. BrowserForge differs in \emph{where
the data comes from}: rather than expanding the number of trajectories within a
fixed site distribution, we widen the site distribution itself by sourcing URLs
from Common Crawl~\citep{commoncrawl}, so that scale and diversity grow together.
The Proposer--Solver loop and the cleaning pipeline are what make unsupervised
open-web pages usable as training data.

\paragraph{Web agents, benchmarks, and multimodal GUI models.}
Early LLM-based web agents operated over HTML or accessibility trees, and the
shift to multimodal backbones let agents act directly on rendered pages.
SeeAct~\citep{zheng2024seeact} showed that a strong LMM can plan web actions
visually when its plans are grounded onto page elements, and
WebVoyager~\citep{he2024webvoyager} pushed toward end-to-end agents that
interact with live sites. Benchmarks have tracked this shift:
Mind2Web~\citep{deng2023mind2web} offers $2{,}350$ crowdsourced tasks over $137$
websites; WebLINX~\citep{lu2024weblinx} targets conversational navigation;
WebShop~\citep{yao2022webshop} provides a large simulated shopping environment;
WebArena~\citep{zhou2023webarena} and its multimodal extension
VisualWebArena~\citep{koh2024visualwebarena} host self-contained sites and score
functional correctness; GAIA~\citep{mialon2023gaia} probes general assistants on
real-world tool-use and browsing; and
Online-Mind2Web~\citep{xue2025illusion} evaluates agents on live websites and
reports that several agents are weaker than earlier numbers suggested. As
evaluation suites these are valuable, but as \emph{training} sources they share a
ceiling: each draws from a fixed and relatively small set of sites. A parallel
line trains end-to-end models that perceive screenshots and emit low-level
actions~\citep{guig2,wu2024osatlas,guisage,tang2025survey,gu2025ui,guilibra}. CogAgent~\citep{hong2024cogagent} and SeeClick~\citep{cheng2024seeclick}
established screenshot-only GUI grounding and navigation~\citep{guig2,cheng2024seeclick,clawgui,lin2024showui,lu2026uir1,lu2025uis1,lu2026uicopilot};
OS-Atlas~\citep{wu2024osatlas} scales cross-platform grounding data;
OmniParser~\citep{lu2024omniparser} parses screenshots into structured elements
to ground actions; UGround~\citep{gou2025uground} learns universal visual
grounding across platforms; Aguvis~\citep{xu2024aguvis} and
ShowUI~\citep{lin2024showui} build pure-vision GUI agents with unified action
spaces; Qwen2.5-VL~\citep{bai2025qwen25vl} and Qwen3-VL~\citep{qwen3vl2025}
provide strong open backbones; and GUI-specialized models such as
UI-TARS~\citep{qin2025uitars} and UI-TARS-2~\citep{wang2025uitars2} push success
rates on screenshot-driven control. Beyond the web,
AndroidControl~\citep{li2024androidcontrol} and
AndroidWorld~\citep{rawles2024androidworld} study data scale and dynamic
evaluation for mobile UI control. These models are largely orthogonal to our
contribution: BrowserForge produces data, and a stronger backbone or training
recipe can be combined with it. In our experiments we fine-tune a compact open
multimodal model on the BrowserForge corpus and measure the change in web-agent
performance, treating the data pipeline, not the backbone, as the variable under
study.

\section{Method}
\label{sec:method}

\subsection{Overview}
\label{sec:method_overview}

\begin{figure}[t]
\centering
\includegraphics[width=\linewidth]{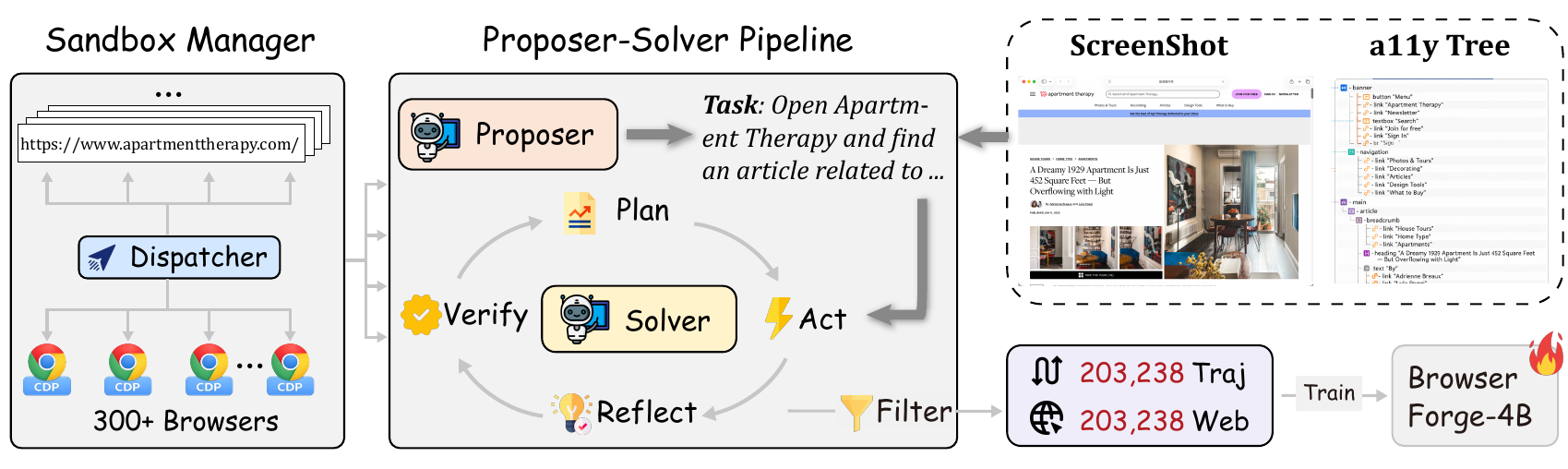
}
\caption{The BrowserForge synthesis framework. A \emph{sandbox cluster manager}
(left) maintains a machine list, a URL work queue, a pool of browser sandboxes,
and worker threads. For each page, the \emph{Proposer} (middle) generates a
candidate task in two stages and selects the most executable one; the
\emph{Solver} (right), built on an open browser-control agent, executes the
task through a plan--act--reflect--verify loop and writes the trajectory to
storage. Verified trajectories are cleaned and formatted into the final
training corpus.}
\label{fig:framework}
\end{figure}

We introcuce \textbf{BrowserForge}, a framework that
turns the open web into web-agent training data, as shown in
Figure~\ref{fig:framework}. The method decomposes into four modules:
  \textbf{(1) Open-web URL sourcing and cleaning}
  (Section~\ref{sec:url}), which draws candidate URLs from the open web and
  filters them so that the agent is placed in front of real, openly reachable
  websites rather than a fixed list;
  \textbf{(2) Parallel browser sandbox orchestration}
  (Section~\ref{sec:sandbox}), a cluster manager that drives hundreds of browser
  sandboxes at once and supplies rendered pages at high throughput;
  \textbf{(3) Proposer--Solver task synthesis}
  (Section~\ref{sec:proposer_solver}), in which a Proposer agent turns a
  rendered page into an executable task and a Solver agent carries it out and
  records the trajectory; and
  \textbf{(4) Trajectory cleaning and unified reasoning}
  (Section~\ref{sec:cleaning}), which filters the collected trajectories and
  rewrites their reasoning into a single format.
Together these modules target three properties at once: \emph{scale}, from
parallel sandboxes; \emph{diversity}, from open-web URL sourcing; and
\emph{quality}, from verification and cleaning.

\subsection{Open-Web URL Sourcing and Cleaning}
\label{sec:url}

Prior trajectory-synthesis methods focus on generating more trajectories, but
they remain bound to the source they start from, whether a curated seed list, a
tutorial corpus, or a fixed environment, which caps the number of distinct
websites the agent ever sees. We instead build the corpus on \emph{real} URLs.
Our sourcing is inspired by Common Crawl~\citep{commoncrawl}, a large-scale,
periodically refreshed snapshot of the public web that spans many millions of
hosts; sampling from such a snapshot rather than a fixed list means the supply
of candidate URLs can, in principle, be scaled up without bound, and it is what
lets BrowserForge reach hundreds of thousands of distinct websites.

Raw Common Crawl URLs, however, are uncleaned: many are dead, toxic, or static
endpoints that afford no interaction. We therefore design a cleaning pipeline
that runs before any sandbox time is spent, combining standard filters with a
reachability check tailored to our own infrastructure:
\begin{itemize}
  \item \textbf{Reachability and type filtering.} We discard URLs that fail to
  resolve or load, and filter by site type to keep pages that afford
  interaction rather than static endpoints (for example, raw media or file
  downloads).
  \item \textbf{Content filtering.} Very short or near-empty pages carry little
  interaction signal and are removed; we keep pages whose main content and
  multimodal elements support a non-trivial task.
  \item \textbf{Blacklist filtering.} We remove URLs that overlap with
  evaluation benchmarks or known sensitive hosts, which keeps the training
  distribution clean and avoids contamination of the test sets used in
  Section~\ref{sec:exp}.
  \item \textbf{IP-level accessibility validation.} A URL that is reachable in
  general may still be blocked from, or fail to render on, the machines that run
  our sandboxes. We therefore re-validate each surviving URL from the cluster's
  own IP addresses and keep only those that remain actually accessible, so that
  no sandbox time is wasted on pages it cannot open.
\end{itemize}
Each surviving URL is rendered inside a sandbox, and the resulting page state is
packaged for the synthesis stage as a tuple of the current screenshot, the page
URL and title, and the accessibility (a11y) tree. To broaden the visual and
locale distribution further, sandboxes are configured with varied screen
resolutions, user agents, and interface languages, which diversifies image
sizes and rendered layouts without changing the underlying task logic.

\subsection{Parallel Browser Sandbox Orchestration}
\label{sec:sandbox}

Scale comes from running many browser sandboxes concurrently, and doing so
without leaving compute idle requires explicit coordination. A \emph{sandbox
cluster manager} owns this fleet and keeps it busy. It maintains four pieces of
state: a machine list of available compute nodes; a shared URL work queue holding the cleaned pages waiting to be processed; a sandbox pool of live browsers, each isolated from the others and
addressable through the Chrome DevTools Protocol (CDP); and a worker
pool of threads, each of which repeatedly claims a free sandbox, pulls the next
URL from the queue, drives the synthesis loop on that page, writes the resulting
trajectory back to storage, and then returns the sandbox to the pool for reuse.

This design scales to up to $300$ browser sandboxes in parallel. The shared
queue is the key to high utilization: because workers pull work rather than
being assigned fixed partitions, a slow or stalled page occupies only its own
worker, and the rest of the fleet keeps draining the queue instead of waiting on
it. Each sandbox can host several independent browser profiles, so a single node
runs multiple concurrent sessions and the cost of browser startup is amortized
across many pages. Compute nodes can be added to the machine list while the
pipeline is running, and the queue rebalances onto them automatically. This
orchestration layer is what turns a single-machine data collector into a
cluster-scale one, and it is the component that most directly determines
throughput.

\subsection{Proposer--Solver Task Synthesis}
\label{sec:proposer_solver}

Open-web sourcing supplies pages, but a rendered page is not yet training data:
it has no task to pursue and no trajectory to learn from. What is missing is a
\emph{task source}. We supply one with a \textbf{Proposer--Solver} framework:
the Proposer turns a page into an executable task, and the Solver carries that
task out and records the trajectory.

\paragraph{Proposer.}
The Proposer reads the packaged page state (the accessibility tree, the
screenshot, and the page URL and title) and produces an executable task in two
stages. In the \emph{proposal} stage, we prompt
Qwen3-VL-235B~\citep{qwen3vl2025} to generate three candidate tasks that are
grounded in the visible page elements and that could plausibly be completed on
this page; the prompt explicitly excludes tasks whose keywords require account
registration, login, or payment, since such tasks cannot be completed
autonomously and are unsafe to attempt. In the \emph{reflection-and-selection}
stage, the model re-examines the three candidates against the page, judges which
are genuinely executable, and selects a single best task. Proposing several
candidates and then selecting among them, rather than committing to the first
idea, filters out under-specified or impossible tasks before any execution
effort is spent and raises the yield of usable trajectories.

\begin{table}[t]
\centering
\setlength{\tabcolsep}{8pt}
\renewcommand{\arraystretch}{1.15}
\resizebox{\linewidth}{!}{%
\begin{tabular}{lll}
\toprule
\textbf{Action} & \textbf{Arguments} & \textbf{Description} \\
\midrule
\texttt{click}     & coordinate              & Click a point on the screen. \\
\texttt{type}      & coordinate, text        & Focus an input field and type text. \\
\texttt{select}    & coordinate, value       & Choose a value from a dropdown / combobox. \\
\texttt{scroll}    & direction, amount       & Scroll the view up/down/left/right by a fraction of the viewport. \\
\texttt{key}       & key                     & Press a single key (e.g.\ Enter, Tab, Escape). \\
\texttt{wait}      & seconds                 & Wait for the page to load or settle. \\
\texttt{go\_back}  & ---                     & Navigate back to the previous page. \\
\texttt{visit\_url}& url                     & Navigate the current tab to an absolute URL. \\
\texttt{finish}    & message                 & Conclude the task and return the answer. \\
\bottomrule
\end{tabular}%
}
\caption{The Solver's unified action space. Screen positions are expressed as
normalized integer coordinates in $[0, 1000]$, independent of the rendered
resolution. At each step the Solver emits exactly one action from this set; the
\texttt{finish} action terminates the trajectory.}
\label{tab:action_space}
\end{table}

\paragraph{Solver.}
Given the selected task, the Solver collects a trajectory by acting on the live
page. It is built on an open browser-control agent~\citep{browseruse2024} and
driven by a multimodal model through a closed loop of \emph{planning},
\emph{acting}, \emph{reflection}, and \emph{verification}, with a running
\emph{memory} of prior steps. At each step the model observes the current page
as a screenshot together with the structured set of interactive elements, and
emits an evaluation of the previous step, an updated memory, the next goal, and
one or more grounded actions (such as clicking an element, typing text, or
scrolling), which are executed in the sandbox. The Solver acts through a unified
action space, summarized in Table~\ref{tab:action_space}, in which screen
positions are expressed as normalized integer coordinates so that the same action
format applies across pages of different resolutions. The verification component
checks
whether the task has been satisfied and decides whether to continue or to stop;
a dedicated terminal action signals completion and ends the run. Because
observation, reasoning, and action are interleaved within a single agent loop,
the Solver can recover from intermediate mistakes rather than committing to a
fixed plan.

\paragraph{From runs to a corpus.}
Each completed run is stored as a structured trajectory, the agent's running
memory together with a per-step JSON record of observations, thoughts, and
actions, and accumulated into the training set. Because the Proposer and Solver
both run inside the sandbox fleet, task generation and trajectory collection
scale with the same parallel infrastructure described in
Section~\ref{sec:sandbox}, so the pipeline produces a large corpus of raw
trajectories at cluster throughput. In total we collect $203{,}238$ raw
trajectories, one per distinct website, averaging $8.8$ steps each, for roughly
$1.8$M interaction steps before any filtering.

\subsection{Trajectory Cleaning and Unified Reasoning Format}
\label{sec:cleaning}

The pipeline so far yields trajectories at scale, but because tasks are proposed
and executed without human supervision, a fraction of the sampled runs fail:
the Solver may pursue a goal that turns out to be impossible, take a broken
action sequence, or stop prematurely, and the recorded reasoning varies widely
in style and length from run to run. Training directly on this raw mixture
passes the noise into the model. We therefore clean the corpus in three steps:
the first two filter the raw collection for correctness, and the third
standardizes the reasoning of the data we ultimately train on.
\begin{itemize}
  \item \textbf{Rule-based filtering.} We check whether the trajectory's final
  action is the terminal \texttt{Finish} action. Runs that never terminate
  properly, or whose action sequences are malformed, are discarded outright,
  removing obvious failures cheaply before any model is invoked.
  \item \textbf{Model-based judging.} For the surviving runs, we prompt
  Qwen3-VL-235B with the task description together with the last three
  screenshots of the trajectory, and ask it to judge whether the task was
  actually accomplished. Trajectories the judge rejects are removed, which
  filters out runs that terminate cleanly but do not in fact complete the task.
  Together these two filters are strict: only about $30\%$ of the raw
  interaction steps survive, leaving roughly $600$K verified steps.
  \item \textbf{Unified chain-of-thought rewriting.} From this verified pool we
  sample $200$K steps and use Seed 2.0 Pro to rewrite their reasoning into a
  single, consistent chain-of-thought format. The Solver's original rationales
  are often overly verbose and heterogeneous; rewriting them gives the
  downstream model a stable supervision target rather than a mix of inconsistent
  reasoning styles. These $200$K Seed-rewritten steps form the final training
  corpus.
\end{itemize}
Overall, BrowserForge sources $203{,}238$ raw trajectories from as many distinct
websites; filtering keeps roughly $30\%$ of their steps, and from this verified
pool we draw the $200$K Seed-rewritten steps used for training.
Table~\ref{tab:dataset_compare} places this corpus against prior web-agent
trajectory datasets along scale, site coverage, and the source of diversity.

\begin{table}[t]
\centering
\setlength{\tabcolsep}{8pt}
\renewcommand{\arraystretch}{1.1}
\resizebox{\linewidth}{!}{%
\begin{tabular}{lcccc}
\toprule
\multirow{2}{*}{\textbf{Dataset}} & \multicolumn{2}{c}{\textbf{Scale}} & \multirow{2}{*}{\textbf{Diversity}} & \multirow{2}{*}{\textbf{Reason}} \\
\cmidrule(lr){2-3}
 & \#Trajectories & \#Websites & & \\
\midrule
Mind2Web~\citep{deng2023mind2web}        & 1{,}009   & 137      & Fixed     & $\times$   \\
WebLINX~\citep{lu2024weblinx}            & 969       & 155      & Fixed     & $\times$   \\
GUIAct~\citep{chen2024guicourse}         & 2{,}482   & 121      & Fixed     & $\times$   \\
OpenWebVoyager~\citep{he2024openwebvoyager} & 1{,}165 & 48     & Fixed     & $\checkmark$   \\
NNetNav~\citep{murty2024nnetnav}         & 10{,}272   & 20        & Fixed     & $\checkmark$   \\
AgentTrek~\citep{xu2025agenttrek}        & 10{,}398  & 127      & Tutorial  & \checkmark \\
Explorer~\citep{pahuja2025explorer}      & 94{,}000  & 49{,}000 & Explore   & \checkmark \\
\midrule
\rowcolor{gray!15}
\textbf{BrowserForge (Ours)} & \textbf{203{,}238} & \textbf{$\sim$200K} & \textbf{Common Crawl} & \checkmark \\
\bottomrule
\end{tabular}%
}
\caption{Comparison of BrowserForge with existing web-agent trajectory datasets.
``Reason'' marks whether task synthesis includes an explicit reasoning/verification
step. Each BrowserForge trajectory is collected from a distinct website, so its
trajectory and website counts coincide; it is the largest in trajectory count and
the broadest in website coverage, sourcing sites from the open web rather than a
fixed list.}
\label{tab:dataset_compare}
\end{table}

\section{Experiments}
\label{sec:exp}

Training a compact multimodal model on the BrowserForge corpus turns it into a
markedly stronger web agent, and controlled experiments trace that gain to the
corpus itself rather than to the training recipe. We organize the evaluation
around this claim: a main comparison on a live benchmark
(Section~\ref{sec:exp_main}), an analysis that isolates the data source and
scale (Section~\ref{sec:exp_analysis}), an ablation of the
synthesis and cleaning pipeline (Section~\ref{sec:ablation}), and an error
analysis of how the trained agent fails (Section~\ref{sec:error_analysis}).

\subsection{Experimental Setup}
\label{sec:exp_setup}

\textbf{Implementation details.}
We fine-tune the compact open Qwen3.5 multimodal models at $4$B and $9$B to
obtain \textbf{BrowserForge-4B} and \textbf{BrowserForge-9B}. Training runs on
the LLaMA-Factory framework~\citep{zheng2024llamafactory} for $3$ epochs at a
learning rate of $1\mathrm{e}{-5}$, with a maximum of $99{,}999{,}999$ pixels per
image so that pages are effectively left uncropped. We freeze the visual encoder
and the adapter (projection) layers and perform full-parameter supervised
fine-tuning of the language-model component only, supervising the assistant
response tokens. Training and inference share one fixed system prompt that
conditions on the current screenshot and emits a single action from the unified
action space per step (Appendix~\ref{sec:appendix_prompt}).

\textbf{Benchmarks and metric.}
We evaluate on two complementary benchmarks. \textit{Online-Mind2Web}
~\citep{xue2025illusion} measures end-to-end task success on $300$ tasks over live
websites; following prior practice, success is scored by the WebJudge-7B
~\citep{xue2025illusion} automatic judge and we report success rate (SR, \%).
\textit{Multimodal-Mind2Web} scores \emph{step-wise} success across its
Cross-Task, Cross-Website, and Cross-Domain splits under a fixed-trajectory
protocol, probing fine-grained grounding rather than full task completion.

\textbf{Baselines.}
We compare against two groups: (i) proprietary GUI agents, shown only as an
upper-reference point rather than as controlled comparisons; and (ii) open
multimodal models, including Qwen3-VL and the Qwen3.5 series evaluated
zero-shot, alongside open-source GUI and web-agent methods.

\subsection{Main Results}
\label{sec:exp_main}

We evaluate on two mainstream web-agent benchmarks, \textit{Online-Mind2Web}, a
dynamic benchmark that measures end-to-end task success on live websites, and
\textit{Multimodal-Mind2Web}, a static benchmark that measures step-wise accuracy
under a fixed-trajectory protocol. As shown in Table~\ref{tab:online_mind2web} and
Table~\ref{tab:multimodal_mind2web}, the results support three findings: 

\textbf{(1) BrowserForge data yields a consistent gain on both the dynamic and the
static benchmark.} On the dynamic Online-Mind2Web, fine-tuning Qwen3.5-4B on the
BrowserForge corpus raises success rate from $25.66\%$ to $33.33\%$ ($+7.67$), and
Qwen3.5-9B from $29.33\%$ to $38.00\%$ ($+9.33$). On the static
Multimodal-Mind2Web, average step accuracy rises in lockstep, from $38.2\%$ to
$43.8\%$ (Pass@1) for the $4$B model and from $40.0\%$ to $45.1\%$ for the $9$B
model. The improvement appears on both protocols and both backbones rather than at
a single setting.

\begin{table}[t]
\centering
\small
\setlength{\tabcolsep}{5.5pt}
\renewcommand{\arraystretch}{1.08}
\begin{tabular*}{\linewidth}{@{\extracolsep{\fill}}llcr@{}}
\toprule
\textbf{Model}
& \textbf{Domain}
& \textbf{Agent Traj.}
& \textbf{SR (\%)} \\
\midrule

\rowcolor{groupgray}
\multicolumn{4}{@{}l}{\textit{Proprietary Models}} \\
Browser-Use
    & Web Agent
    & N/A
    & 30.0 \\
Gemini-CUA
    & Computer-Use Agent
    & N/A
    & 69.0 \\
Navigator
    & Web Agent
    & N/A
    & 78.7 \\
Operator
    & Computer-Use Agent
    & Undisc.
    & 61.3 \\
GPT-5 + UGround-v1-7B
    & Planner \& Grounder
    & N/A
    & 33.3 \\

\addlinespace[2pt]
\rowcolor{groupgray}
\multicolumn{4}{@{}l}{\textit{Base MLLMs}} \\
Qwen2.5-VL-3B
    & General VLM
    & N/A
    & 8.3 \\
Qwen2.5-VL-7B
    & General VLM
    & N/A
    & 22.0 \\
Qwen2.5-VL-32B
    & General VLM
    & N/A
    & 19.7 \\
Qwen3-VL-4B
    & General VLM
    & N/A
    & 27.7 \\
Qwen3-VL-8B
    & General VLM
    & N/A
    & 27.7 \\
Qwen3-VL-32B
    & General VLM
    & N/A
    & 34.3 \\

\addlinespace[2pt]
\rowcolor{groupgray}
\multicolumn{4}{@{}l}{\textit{Existing Methods}} \\
ScaleCUA-3B
    & Computer-Use Agent
    & 19K
    & 21.7 \\
ScaleCUA-7B
    & Computer-Use Agent
    & 19K
    & 30.3 \\
ScaleCUA-32B
    & Computer-Use Agent
    & 19K
    & 30.0 \\
GUI-Libra-3B
    & Web Agent
    & 9K
    & 29.0 \\
GUI-Libra-4B
    & Web Agent
    & 9K
    & 31.3 \\
GUI-Libra-7B
    & Web Agent
    & 9K
    & 33.3 \\
GUI-Libra-8B
    & Web Agent
    & 9K
    & \underline{36.7} \\
WebStar-7B
    & Web Agent
    & 13.3K
    & 22.8 \\
WebStar-32B
    & Web Agent
    & 13.3K
    & 23.8 \\

\addlinespace[2pt]
\rowcolor{groupgray}
\multicolumn{4}{@{}l}{\textit{Ours}} \\
Qwen3.5-4B (baseline)
    & General VLM
    & N/A
    & 25.7 \\
\rowcolor{oursblue}
\textbf{BrowserForge-4B}
    & Web Agent
    & \textbf{20K}
    & 33.3\,\gain{7.6} \\
Qwen3.5-9B (baseline)
    & General VLM
    & N/A
    & 29.3 \\
\rowcolor{oursblue}
\textbf{BrowserForge-9B}
    & Web Agent
    & \textbf{20K}
    & \textbf{38.0}\,\gain{9.3} \\
\bottomrule
\end{tabular*}

\caption{Success rate (SR, \%) on Online-Mind2Web, scored using the
WebJudge-7B automatic judge. \textit{Agent Traj.}
denotes the reported number of agent-interaction trajectories used for
specialized training. ``Undisc.'' indicates that the training scale is not
publicly disclosed, while ``--'' indicates that no specialized agent
trajectories are used. Proprietary models are included for reference only.
\textbf{Bold} denotes the best result among the compared base models, existing
methods, and our models; \underline{underlining} denotes the best prior result.
BrowserForge-4B/9B are fine-tuned from Qwen3.5-4B/9B on the BrowserForge
corpus, with absolute gains over the corresponding backbones shown in green.}
\label{tab:online_mind2web}
\end{table}

\textbf{(2) A compact BrowserForge agent surpasses much larger open-source models
and competitive proprietary ones.} On Online-Mind2Web, BrowserForge-4B already
overtakes larger open-source agents such as ScaleCUA-7B ($30.3\%$) and
GUI-Libra-4B ($31.3\%$), and BrowserForge-9B at $38.0\%$ exceeds the strongest
prior open-source method GUI-Libra-8B ($36.7\%$). Notably, GUI-Libra is trained
with reinforcement learning, whereas BrowserForge uses supervised fine-tuning
only, so the gain comes from the data rather than a heavier training recipe. The
$9$B model also clears strong proprietary references, including GPT-5 with UGround
($33.3\%$) and Browser-Use ($30.0\%$). That a $4$B/$9$B model reaches this level
points to data quality, not parameter count, as the driver.

\textbf{(3) The gain is robust across distribution shifts.} On Multimodal-Mind2Web
(Table~\ref{tab:multimodal_mind2web}), the improvement holds on every split,
Cross-Task, Cross-Website, and Cross-Domain, rather than coming from one
favorable distribution. BrowserForge-4B improves Pass@4 average step accuracy over
its backbone from $44.1\%$ to $54.4\%$ and BrowserForge-9B from $47.2\%$ to
$56.3\%$, with the scaling trend shown in Figure~\ref{fig:analysis}(a).

In summary, training on BrowserForge data with supervised fine-tuning alone turns
a compact backbone into a web agent that improves consistently across dynamic and
static benchmarks and across distribution shifts, reaching a level competitive
with far larger open-source and proprietary systems.

\begin{table}[t]
\centering
\small
\setlength{\tabcolsep}{4pt}
\renewcommand{\arraystretch}{1.1}
\resizebox{\linewidth}{!}{%
\begin{tabular}{lcccccccc}
\toprule
\multirow{2}{*}{\textbf{Model}} & \multicolumn{2}{c}{\textbf{Cross-Task}} & \multicolumn{2}{c}{\textbf{Cross-Website}} & \multicolumn{2}{c}{\textbf{Cross-Domain}} & \multicolumn{2}{c}{\textbf{Average}} \\
\cmidrule(lr){2-3}\cmidrule(lr){4-5}\cmidrule(lr){6-7}\cmidrule(lr){8-9}
 & Pass@1 & Pass@4 & Pass@1 & Pass@4 & Pass@1 & Pass@4 & Pass@1 & Pass@4 \\
\midrule
\rowcolor{groupgray}
\multicolumn{9}{l}{\textit{Proprietary Models with SeeAct-V Framework}} \\
\midrule
GPT-4o + UGround-v1-7B        & 35.7 & 38.9 & 33.9 & 37.6 & 39.1 & 42.2 & 36.2 & 39.6 \\
GPT-4.1 + UGround-v1-7B       & 41.1 & 44.8 & 36.2 & 39.7 & 43.0 & 46.4 & 40.1 & 43.6 \\
GPT-5-mini + UGround-v1-7B    & 44.2 & 48.0 & 40.4 & 44.2 & 45.8 & 48.1 & 43.5 & 46.7 \\
GPT-5 + UGround-v1-7B         & 47.7 & 51.7 & 45.0 & 47.6 & 48.2 & 51.6 & 47.0 & 50.3 \\
\midrule
\rowcolor{groupgray}
\multicolumn{9}{l}{\textit{Open-source Native Models}} \\
\midrule
GUI-R1-3B               & 24.0 & 37.7 & 22.3 & 37.7 & 24.6 & 39.1 & 23.6 & 38.2 \\
GUI-R1-7B               & 37.0 & 50.1 & 34.1 & 46.3 & 39.6 & 50.5 & 36.9 & 49.0 \\
Aguvis-7B               & 37.7 & 48.0 & 31.7 & 41.5 & 36.9 & 45.1 & 35.4 & 44.9 \\
UI-TARS-1.5-7B          & 37.2 & 48.0 & 31.3 & 42.9 & 35.6 & 47.2 & 34.7 & 46.0 \\
GLM-4.1V-9B-Thinking    & 26.9 & 32.9 & 23.0 & 29.3 & 28.7 & 35.3 & 26.2 & 32.5 \\
Qwen2.5-VL-3B           & 24.4 & 29.0 & 18.6 & 24.6 & 27.1 & 31.2 & 23.4 & 28.3 \\
Qwen2.5-VL-7B           & 31.6 & 45.6 & 30.4 & 42.1 & 35.6 & 48.0 & 32.5 & 45.2 \\
Qwen2.5-VL-32B          & 46.2 & 55.8 & 42.6 & \textbf{55.7} & 46.0 & 57.9 & 44.9 & 56.5 \\
Qwen3-VL-4B             & 43.5 & 49.9 & 39.1 & 48.7 & 43.2 & 50.1 & 41.9 & 49.6 \\
Qwen3-VL-8B             & 45.9 & 52.8 & 40.0 & 48.3 & 41.8 & 52.3 & 42.6 & 51.1 \\
\midrule
Qwen3.5-4B              & 39.5 & 45.1 & 35.1 & 40.8 & 40.0 & 46.3 & 38.2 & 44.1 \\
Qwen3.5-9B              & 42.1 & 48.3 & 36.5 & 43.6 & 41.5 & 49.7 & 40.0 & 47.2 \\
\midrule
\rowcolor{oursblue}
\textbf{BrowserForge-4B (Ours)} & 47.0 & 56.8 & 41.5 & 52.4 & 42.9 & 54.0 & 43.8 & 54.4 \\
\rowcolor{oursblue}
\textbf{BrowserForge-9B (Ours)} & 48.1 & 57.9 & 41.3 & 54.6 & 45.9 & 56.5 & 45.1 & 56.3 \\
\bottomrule
\end{tabular}%
}
\caption{Step accuracy (\%) on Multimodal-Mind2Web. Proprietary models are shown
for reference. \textbf{Bold}: best; \underline{underline}: second best among
open-source methods. BrowserForge-4B/9B are fine-tuned from Qwen3.5-4B/9B on the
BrowserForge corpus and improve over their respective backbones on every split.}
\label{tab:multimodal_mind2web}
\end{table}

\subsection{Analysis}
\label{sec:exp_analysis}

The main results establish that BrowserForge data helps; the central claim of
this paper, however, is about \emph{where} the data comes from. We therefore run
two controlled experiments that isolate the data source and its scale from the
model and optimizer.

\textbf{Source: the open-web corpus, not the training recipe, drives the gain.}
A natural concern is that any large fine-tuning set would help equally well. To
rule this out, we hold the Qwen3.5-4B backbone and the $3$-epoch budget fixed and
vary only the data, fine-tuning on existing open-source web trajectories versus
the BrowserForge corpus. For a fair comparison, we sample $200$K trajectories
from the open-source data to match the size of our training set. As shown in
Table~\ref{tab:data_compare}, training on
BrowserForge data raises average step accuracy on Multimodal-Mind2Web from
$32.74\%$ to $41.33\%$ (Pass@1) and from $43.48\%$ to $52.59\%$ (Pass@4), a
gain of close to nine percent that holds across all three splits. Because
everything but the data is identical, the gain is attributable to the corpus, to
its open-web diversity and the rule-plus-model cleaning, rather than to the
architecture or optimizer.

\begin{table}[t]
\centering
\small
\setlength{\tabcolsep}{4pt}
\renewcommand{\arraystretch}{1.1}
\resizebox{\linewidth}{!}{%
\begin{tabular}{lcccccccc}
\toprule
\multirow{2}{*}{\textbf{Training data}} & \multicolumn{2}{c}{\textbf{Cross-Task}} & \multicolumn{2}{c}{\textbf{Cross-Website}} & \multicolumn{2}{c}{\textbf{Cross-Domain}} & \multicolumn{2}{c}{\textbf{Average}} \\
\cmidrule(lr){2-3}\cmidrule(lr){4-5}\cmidrule(lr){6-7}\cmidrule(lr){8-9}
 & Pass@1 & Pass@4 & Pass@1 & Pass@4 & Pass@1 & Pass@4 & Pass@1 & Pass@4 \\
\midrule
Open-source trajectories        & 36.45 & 45.41 & 28.85 & 41.02 & 32.93 & 44.01 & 32.74 & 43.48 \\
\rowcolor{oursblue}
\textbf{BrowserForge (Ours)}     & 44.88 & 54.74 & 37.88 & 50.83 & 41.22 & 52.20 & 41.33 & 52.59 \\
\bottomrule
\end{tabular}%
}
\caption{Controlled comparison of the training-data source on Multimodal-Mind2Web
step accuracy (\%). Holding the Qwen3.5-4B backbone and the $3$-epoch training
budget fixed, we fine-tune on existing open-source web trajectories versus the
BrowserForge corpus. Read across the two rows: the gap isolates the effect of the
data source rather than the architecture or optimizer.}
\label{tab:data_compare}
\end{table}

\begin{figure}[t]
\centering
\includegraphics[width=\linewidth]{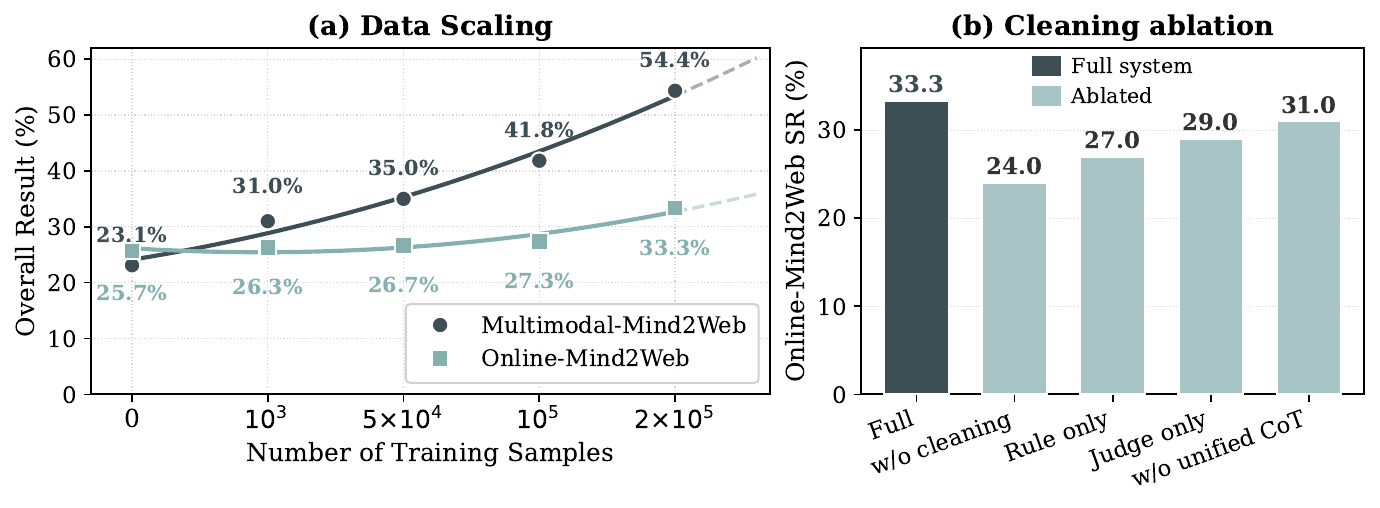}
\caption{\textbf{(a) Data scaling.} Overall result versus the number of
BrowserForge training samples (log scale), fine-tuning Qwen3.5-4B for $3$ epochs
under a fixed budget, on both Multimodal-Mind2Web (step accuracy) and
Online-Mind2Web (success rate). The $0$-sample point is the zero-shot base model;
solid curves are power-law fits over the measured points and dashed segments
extrapolate beyond the largest run. \textbf{(b) Cleaning ablation.}
Online-Mind2Web success rate for each variant of the cleaning pipeline
(rule-based filter, model-based judge, and Seed-rewritten unified
chain-of-thought), with the full system highlighted.}
\label{fig:analysis}
\end{figure}

\textbf{Scale: performance grows monotonically with the amount of BrowserForge
data.} If open-web sourcing is the right lever, more of it should yield a better
agent. We fine-tune the same backbone on nested subsets of the $\sim$$200$K
unified training set ($10^3$, $5{\times}10^4$, $10^5$, and the full
$2{\times}10^5$ samples) under an identical $3$-epoch budget, against the
zero-shot base model ($0$ samples). Figure~\ref{fig:analysis}(a) plots the trend
on both benchmarks. Performance rises monotonically throughout: Online-Mind2Web
success climbs from $25.66\%$ at zero-shot to $27.33\%$ at $10^5$ and $33.33\%$ at
the full set, while Multimodal-Mind2Web step accuracy rises from $23.11\%$ to
$41.83\%$ and $54.36\%$ over the same range. The curve follows an approximately
power-law shape with no sign of saturating at $2{\times}10^5$, supporting the
premise that open-web sourcing keeps improving the agent as the corpus grows.

\subsection{Ablation}
\label{sec:ablation}

The cleaning pipeline is what turns the raw, noisy trajectories into verified
supervision with a consistent reasoning format, so we ask how much each of its
stages actually contributes to the
trained agent. Starting from the full system, which applies the rule-based filter,
the model-based judge, and the Seed-rewritten unified chain-of-thought in
sequence, we construct four reduced variants that each train the same Qwen3.5-4B for the same $3$ epochs and differ only in how the data is cleaned:
(i) \emph{w/o cleaning}, trained on the raw trajectories with no filtering or
rewriting; (ii) \emph{rule-based filter only}, which keeps trajectories that
terminate with a valid \texttt{Finish} action but applies neither the judge nor
the rewriting; (iii) \emph{model judge only}, which keeps trajectories the judge
accepts but skips the rule filter and the rewriting; and (iv) \emph{w/o unified
CoT}, which applies both filters but trains on the Solver's original reasoning
instead of the unified format. All variants are evaluated by success rate on
Online-Mind2Web under the identical protocol used above.

As shown in Figure~\ref{fig:analysis}(b), we draw three findings:
\textbf{(1) Verification is essential.} Removing cleaning entirely and training on
raw trajectories gives the weakest agent ($24.0\%$), well below the full system
($33.3\%$); the noise from failed and incomplete runs passes directly into the
model. \textbf{(2) The two filters are complementary.} Applying only the rule
filter ($27.0\%$) or only the model judge ($29.0\%$) each recovers part of the gap
but neither matches using both, since the rule filter removes malformed runs
cheaply while the judge catches runs that terminate cleanly yet do not complete the
task. \textbf{(3) A unified reasoning format adds a further gain.} Retaining both
filters but dropping the Seed-rewritten chain-of-thought ($31.0\%$) trails the full
system, showing that a consistent supervision target, not just correct
trajectories, helps the model learn. Overall, every stage of the pipeline
contributes, and the full combination of filtering and unified-CoT rewriting yields
the strongest agent at $33.3\%$.

\subsection{Error Analysis}
\label{sec:error_analysis}

\textbf{Most failures are not random: the agent gets stuck in unproductive
repetition, pointing to a missing feedback signal rather than a missing skill.}
We sample $100$ Online-Mind2Web trajectories of BrowserForge-4B and bucket each
into a single outcome with a priority rule. Only $11\%$ terminate cleanly; the
rest fall into a few recurring patterns, as shown in
Table~\ref{tab:failure_modes}. (1) \textbf{Click loop ($29\%$)} is a perception
gap: the agent clicks the same coordinate four or more times while the page does
not change, mis-targeting an element and failing to notice the action had no
effect, so it does not ground the consequence of an action in the next
observation and repeats the identical click instead of switching strategies.
(2) \textbf{Aimless back-navigation ($22\%$ and $20\%$)} shows the same pattern in
navigation: \texttt{GoBack} signals self-monitoring, but looped it degrades into
ineffective exploration, with $22\%$ of runs issuing $3+$ \texttt{GoBack} actions
and $20\%$ ending in $5+$ consecutive \texttt{Scroll} actions, oscillating between
the two without committing to a concrete sub-goal. (3) \textbf{Access blocking
($9\%$)} is an environment limit: the reasoning explicitly mentions an access wall
(``access denied'', a Cloudflare interstitial, or a CAPTCHA), which the judge
counts as a failure even though the cause is external and the agent correctly
perceives the block, so the gap is the inability to act on it rather than to
recognize it. (4) \textbf{Step-limit timeout ($8\%$)} occurs when a run exhausts
the $30$-step budget while still making plausible progress, so the cap is a partly
artificial source of failure that makes the reported success rate a conservative
lower bound.

\begin{table}[t]
\centering
\setlength{\tabcolsep}{8pt}
\renewcommand{\arraystretch}{1.15}
\begin{tabular}{llc}
\toprule
\textbf{Outcome} & \textbf{Description} & \textbf{\%} \\
\midrule
\rowcolor{gray!15}
Task completed   & Terminates with \texttt{Finish}/\texttt{Exit} or a final answer & 11.0 \\
\midrule
Click loop       & Same coordinate clicked $\geq 4$ times with no state change & 29.0 \\
Back loop        & $\geq 3$ \texttt{GoBack} actions (often paired with scrolling) & 22.0 \\
Scroll-only tail & Run ends with $\geq 5$ consecutive \texttt{Scroll} actions & 20.0 \\
Access blocked   & Reasoning mentions access denied / CAPTCHA / bot wall & 9.0 \\
Step limit       & Exhausts the $30$-step budget without an above pattern & 8.0 \\
No action        & Empty trajectory (no action emitted) & 1.0 \\
\bottomrule
\end{tabular}
\caption{Outcome distribution over $100$ BrowserForge-4B trajectories sampled
from Online-Mind2Web. The top row is the only success category; the remaining
rows are mutually exclusive failure modes, assigned by a priority rule and
ordered by frequency. The three dominant failures, namely repeated clicking,
repeated back-navigation, and trailing scrolls, are all forms of \emph{unproductive
repetition} that together account for $71\%$ of trajectories.}
\label{tab:failure_modes}
\end{table}

\section{Conclusion}
\label{sec:conclusion}

We presented BrowserForge, a framework that generates web episodes at
scale by driving hundreds of browser sandboxes in parallel over openly sourced
URLs. By drawing candidate sites from the open web, scheduling sandboxes through
a shared work queue, and pairing a Proposer that writes executable tasks with a
Solver that collects verified trajectories, BrowserForge decouples the scale and
diversity of web-agent data from any fixed website list. A rule-plus-model
cleaning pipeline and a unified reasoning format turn the raw, noisy trajectories
into verified training data with a consistent supervision target. The resulting corpus of $203{,}238$ trajectories, each from a
distinct website, is larger and broader than prior trajectory datasets.
Fine-tuning a compact multimodal model on it raises success rate on the live
Online-Mind2Web benchmark from $25.66\%$ to $33.33\%$ and improves step accuracy
on the static Multimodal-Mind2Web benchmark, and controlled analyses trace the
gain to the open-web data source and its website diversity rather than to the
training recipe alone.

\bibliography{iclr2026_conference}
\bibliographystyle{iclr2026_conference}

\appendix
\section{Appendix}
\label{sec:appendix}
Additional implementation details, dataset examples, and qualitative
trajectories will be provided here.

\subsection{Training and Inference System Prompt}
\label{sec:appendix_prompt}
We fine-tune and evaluate the model with a single fixed system prompt that, at
each step, conditions on the current page and emits the next action from the
unified action space of Table~\ref{tab:action_space}. Using the same prompt for
both \emph{training} the model on the BrowserForge corpus and \emph{evaluating}
the fine-tuned model at inference time keeps the input format the model is
trained on identical to the one it is tested on. The prompt fixes the coordinate
convention (normalized integers in $[0, 1000]$), enumerates the available
actions and their arguments, and specifies the exact three-line output format
(\texttt{Thought} / \texttt{Description} / \texttt{Action}). At run time the
trailing \texttt{\{instruction\}} and \texttt{\{history\}} placeholders are
filled with the current task and the sequence of previous actions, and the
current screenshot is supplied as the image input. The full prompt is shown
below.

\begin{lstlisting}[style=promptbox]
You are a browser GUI agent. Given an instruction, the current screenshot, and the history of previous actions, predict the SINGLE next action that best progresses toward completing the user's instruction.

Coordinates are normalized to the [0, 1000] range. Use integers in this range regardless of the actual image resolution.

### Action Space

- click(coordinate): Click on a point on the screen. `coordinate` is `[x, y]` with integers in [0, 1000].
- type(coordinate, text): Click on the input field at `coordinate` [x, y] to focus it, then type `text`.
- select(coordinate, value): Select a value from a dropdown/combobox. `coordinate` is `[x, y]` of the dropdown element, `value` is the option text to select.
- scroll(direction, amount): Scroll the visible page in the given direction by `amount`. `direction` is one of "up", "down", "left", "right" (screen-relative; "down" reveals content below the current view). `amount` is a float in [0.05, 1.0] denoting the fraction of the viewport, e.g. 0.5 = half a screen.
- key(key): Press a single keyboard key, e.g. "Enter", "Tab", "Escape", "Backspace".
- wait(seconds): Wait for the page to load or settle. `seconds` is a positive integer.
- go_back(): Navigate back to the previous page in the current tab.
- visit_url(url): Navigate the current tab to the given absolute URL.
- finish(message): Conclude the task and reply to the user. Use the prefix "[ANSWER] ..." to give the requested answer. Use exactly "[EXIT]" as the message if the answer was already provided in a previous step and you only need to terminate.

### Output Format

Output exactly the following three lines, in this order, with no extra text before or after:

Thought: your reasoning about the current screenshot and the next step
Description: a short natural-language description of the action
Action: a single-line JSON object of the form {"action": "action_name", ...args}, where the fields match the action signature above.

### Instruction
{instruction}

### History
{history}
\end{lstlisting}

\end{document}